\documentclass[letterpaper, 10 pt, conference]{ieeeconf}  

\IEEEoverridecommandlockouts                              

\usepackage{graphics} 
\usepackage{epsfig} 
\usepackage{mathptmx} 
\usepackage{times} 
\usepackage{amsmath} 
\usepackage{amssymb}  
\usepackage{booktabs}

\title{\LARGE \bf
UAV-Based Intelligent Traffic Surveillance System: Real-Time Vehicle Detection, Classification, Tracking, and Behavioral Analysis
}

\author{Ali Khanpour$^{1\dag}$, Tianyi Wang$^{1}$, Afra Vahidi-Shams$^{2}$, Wim Ectors$^{3}$, Farzam Nakhaie$^{4}$, \\
Amirhossein Taheri$^{5}$, Christian Claudel$^{1}$%
\thanks{$\dag$Corresponding author: Ali Khanpour.}%
\thanks{$^{1}$Ali Khanpour, Tianyi Wang, and Christian Claudel are with the Department of Civil, Architectural, and Environmental Engineering, University of Texas at Austin, Austin, TX 78712, USA. Email: ali.khanpour@utexas.edu; bonny.wang@utexas.edu; christian.claudel@utexas.edu.}%
\thanks{$^{2}$Afra Vahidi-Shams is with the Faculty of Electrical and Computer Engineering, Babol Noshirvani University of Technology, Babol, Iran. Email: afra@vahidishams.ir.}%
\thanks{$^{3}$Wim Ectors is with the Transportation Research Institute (IMOB), UHasselt-Hasselt University, Agoralaan, Diepenbeek 3590, Belgium. Email: wim.ectors@uhasselt.be.}%
\thanks{$^{4}$Farzam Nakhaie is with the Faculty of Mechanical Engineering, Babol Noshirvani University of Technology, Babol, Iran. Email: farzamnkh75@gmail.com.}%
\thanks{$^{5}$Amirhossein Taheri is with the Department of Computer Engineering, Amirkabir University of Technology, Tehran, Iran. Email: ah\_taheri@aut.ac.ir.}%
}

\begin{document}

\maketitle
\thispagestyle{empty}
\pagestyle{empty}

\begin{abstract}

Traffic congestion and violations pose significant challenges for urban mobility and road safety. 
Traditional traffic monitoring systems, such as fixed cameras and sensor-based methods, are often constrained by limited coverage, low adaptability, and poor scalability. 
To address these challenges, this paper introduces an advanced unmanned aerial vehicle (UAV)-based traffic surveillance system capable of accurate vehicle detection, classification, tracking, and behavioral analysis in real-world, unconstrained urban environments. 
The system leverages multi-scale and multi-angle template matching, Kalman filtering, and homography-based calibration to process aerial video data collected from altitudes of approximately 200 meters. 
A case study in urban area demonstrates robust performance of the system, achieving a detection precision of 91.8 \%, an F1-score of 90.5 \%, and tracking metrics (MOTA/MOTP) of 92.1 \% and 93.7 \%, respectively. 
Beyond precise detection, the system classifies five vehicle types and automatically detects critical traffic violations, including unsafe lane changes, illegal double parking, and crosswalk obstructions, through the fusion of geofencing, motion filtering, and trajectory deviation analysis. 
The integrated analytics module supports origin–destination tracking, vehicle count visualization, inter-class correlation analysis, and heatmap-based congestion modeling. 
Additionally, the system enables entry–exit trajectory profiling, vehicle density estimation across road segments, and movement direction logging, supporting comprehensive multi-scale urban mobility analytics. 
Experimental results confirms the system's scalability, accuracy, and practical relevance, highlighting its potential as an enforcement-aware, infrastructure-independent traffic monitoring solution for next-generation smart cities.

\end{abstract}

\begin{keywords}
UAV traffic monitoring; vehicle detection; origin-destination analysis; computer vision; smart mobility. 
\end{keywords}

\section{Introduction}
\label{sec:0}

In recent years, the development of unmanned aerial vehicles (UAVs) has accelerated, promising to transform urban mobility and intelligent transportation systems \cite{khan2017uav,outay2020applications}. 
UAVs are increasingly used in traffic monitoring and management, which are critical for ensuring smooth traffic flow, minimizing congestion, and improving safety \cite{butilua2022urban}. 
Large-scale fixed camera infrastructures such as the I-24 MOTION project in Tennessee~\cite{gloudemans202324} have demonstrated the capability to collect continuous high resolution vehicle trajectories over freeway segments using hundreds of synchronized HD cameras.
Collection of detailed traffic data with conventional monitoring methods, including fixed video cameras and inductive loop sensors, suffer from limited coverage and high maintenance costs \cite{morris2017intersection}, while UAVs are capable of providing high-resolution data with the advantages of low costs, high flexibility, and wide monitoring range \cite{barmpounakis2016unmanned}. 
In order to accelerate the application of UAV technologies in traffic monitoring and management, this study proposes an intelligent UAV-based traffic surveillance system that utilizes advanced computer vision algorithms to detect, track, and analyze vehicle movements in real-time.

Recent studies have been attempting to extract various traffic parameters and vehicle trajectories in an automatic environment by using state-of–the-art object detection and tracking algorithms \cite{sharma2021video,avcsar2022moving}.
Intersection of union (IoU) is one of the crucial parameters to track detected objects. 
Huang et al. \cite{huang2017enabling} calculated the IoU of currently detected objects with the predicted bounding boxes of previously detected objects to identify the objects in different frames. 
Hua et al. \cite{hua2018vehicle} designed an efficient algorithm to define initial object tracks based solely on the overlap of detected object bounding boxes in consecutive frames with the IoU scores.
Yang et al. \cite{yang2020vehicle} proposed a framework consisting of a you only look once (YOLO) postprocessing-based detector and a deep learning-based tracker.
A Kalman filter is also used to eliminate the occlusion between vehicles at the tracking process. 
Cao et al. \cite{cao2013tracking} proposed a tracking-by-prediction method, where a Kalman filter was used after measuring the velocity similarity between the vehicles to track them in different groups.
Jiao and Wang \cite{jiao2022traffic} investigated the performance of a Kalman filter under the circumstance of background occlusions, compared with an IoU-based tracker.

\begin{figure*}[ht]
    \centering
    \includegraphics[width=\textwidth]{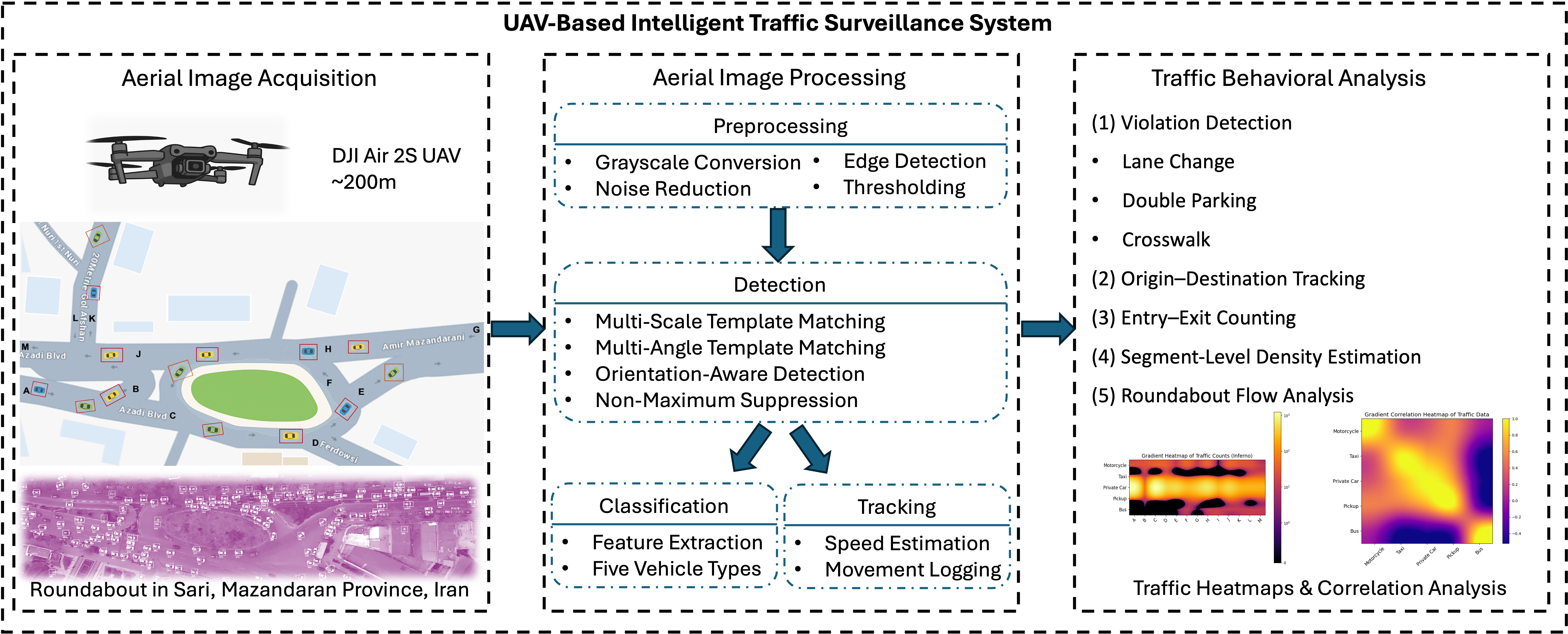}
    \caption{Framework of the proposed UAV-based traffic monitoring system:
    \textbf{(1) Aerial Image Acquisition:} UAVs capture real-time, high-resolution imagery from an optimal altitude, ensuring wide coverage of road networks and granularity needed for reliable vehicle detection.
    \textbf{(2) Computer Vision-Based Processing:} The captured video frames are subjected to a series of preprocessing steps to enhance the image quality. This is followed by deep learning-based object detection to accurately extract vehicle information.
    \textbf{(3) Traffic Data Analysis and Visualization:} Extracted vehicle attributes (e.g., speed, position, and direction) are analyzed and visualized for traffic monitoring, incident detection and decision-making purposes.}
    \label{fig1:framework}
\end{figure*}

Existing studies have also employed UAV-based traffic data to analyze the traffic flow conditions at signalized intersections.
Pitre et al. \cite{pitre2012uav} used UAVs for joint search and track missions, and integrated object detection, tracking, and survivability into a single optimization objective, which was expected to obtain the most valuable information.
Barmpounakis et al. \cite{barmpounakis2019accurate} examined the potential of using UAVs as a way of extracting naturalistic trajectory data from aerial video footage from a low volume four-way intersection.
Liu et al. \cite{liu2019real,liu2022real} coordinated the UAV paths and minimized the monitoring cost for road traffic surveillance at intersections. 
Zhang and Hu \cite{zhang2024tracking} designed a statistics-based adaptive traffic control system in the normal intersection scenes.

Although commercial solutions such as DataFromSky and research datasets like VisDrone and UAVDT have demonstrated vehicle type classification capabilities, many UAV-based traffic studies still focus primarily on detection or trajectory extraction without addressing fine-grained classification under real-world aerial conditions. 
In particular, robust multi-class classification from nadir UAV footage remains challenging due to factors like high-altitude resolution loss, occlusion, and illumination variance especially when real-time analysis is desired. 
Besides, most of current UAV-based traffic surveillance studies primarily focus on isolated scenario, e.g. signalized intersections, often overlooking their spatial inter-dependencies with nearby scenarios.

In this paper, the main focus is on the traffic flow analysis of vehicle trajectories acquired via small rotary-wing UAV footage in intersections, roundabouts, and urban road networks. 
With the help of a case study, this paper attempts to realize real-time vehicle detection and tracking, and comprehensive traffic flow analysis using the presented analytical methodology.
This type of analysis conducted on UAV-based high-resolution traffic data may serve as a benchmark for further research into practical applications of UAV-based traffic surveillance systems.

In summary, this paper makes the following contributions:
\begin{itemize}
    \item  We present a modular UAV-based traffic surveillance framework that enables accurate vehicle detection, tracking, and classification using multi-scale template matching, Kalman filtering, and cross-view analysis.
    \item We propose and evaluate a homography-based method for estimating vehicle speeds using aerial video footage, detecting key illegal driving behaviors, and quantifying their spatiotemporal impacts on traffic congestion and pedestrian safety.
    \item A rich traffic data analytics layer enables OD tracking, traffic demand analysis, and heatmap-based visualizations for flow modeling and congestion hotspot detection.
\end{itemize}

This paper is organized as follows: 
Section~\ref{sec:1} reviews related work on UAVs in vehicle detection and tracking, and traffic flow analysis. 
Section~\ref{sec:2} describes the methodology of our traffic surveillance system, including aerial image acquisition, computer vision-based processing, and traffic analysis and visualization. 
Section~\ref{sec:3} presents the experimental results, demonstrating the effectiveness of our approach. 
Finally, Section~\ref{sec:4} concludes the paper and discusses potential future work directions.

\section{Related Work}
\label{sec:1}

\begin{figure*}[ht]
    \centering
    \includegraphics[width=0.8\textwidth]{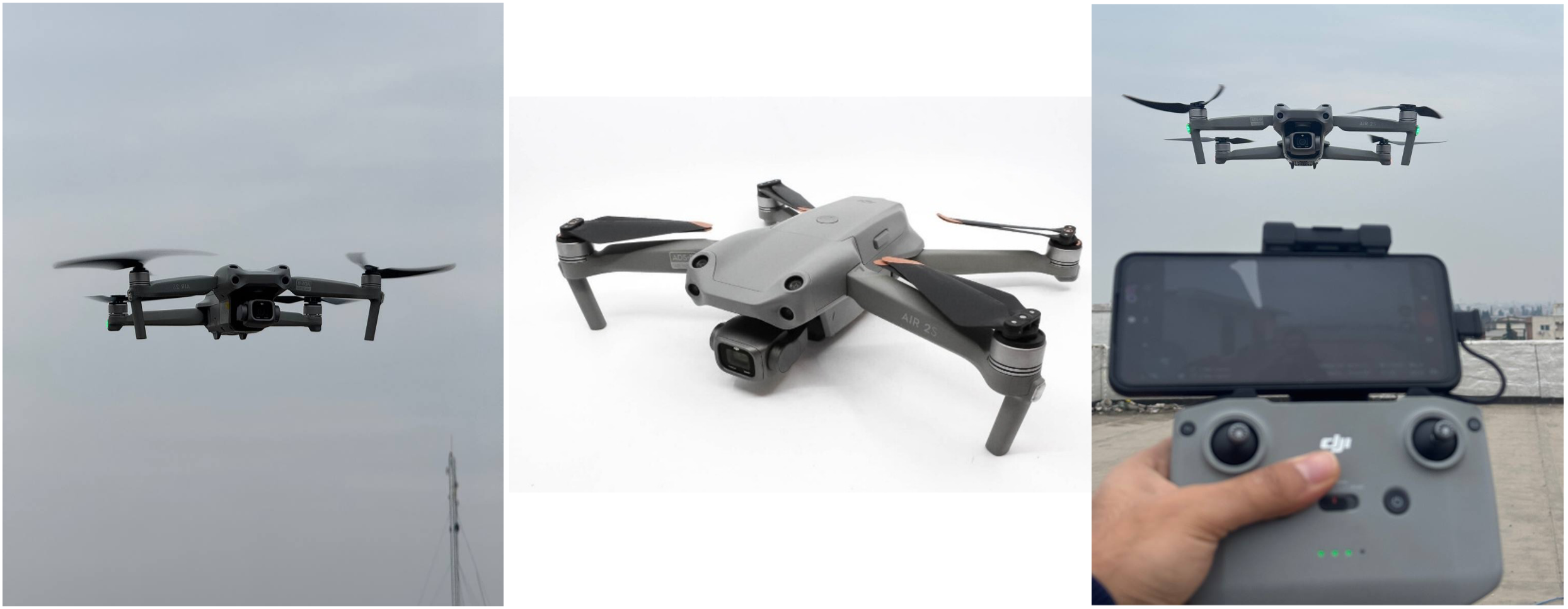}
    \caption{DJI Air 2S UAV utilized for traffic data collection: (left) in-flight operation, (center) pre-flight setup, and (right) live-feed display.}
    \label{fig2:DJIUAV}
\end{figure*}

This section reviews recent advancements in UAV-based traffic surveillance focusing on vehicle detection and tracking, and traffic flow analysis.

\subsection{Unmanned Aerial Vehicles in Vehicle Detection and Tracking}

Vehicle detection and tracking are the initial process for traffic flow analysis \cite{jiang2016surveillance}.
Thus, developing vehicle detectors and trackers with high efficiency, accuracy, and robustness is very important for advanced traffic surveillance tasks \cite{rakai2022data}.
UAVs have proven particularly effective as video data collection tools, offering substantial benefits in extracting vehicle trajectories \cite{kalantar2017multiple}.
The optical flow of targets and backgrounds are also used to predict vehicle location \cite{ke2020advanced}.
Khan et al. \cite{khan2017unmanned,khan2018unmanned1,khan2018unmanned2} proposed a detailed methodological framework for automated UAV traffic video processing and vehicle trajectory extraction, and conducted detailed application and pilot studies of the methodology at urban roundabouts and signalized intersections.
The most well-known and widely used video-based trajectory datasets are NGSIM, HighD, InD, and Interaction. 
In addition, several drone trajectory datasets are mainly used for vehicle detection and classification purposes, utilizing labeled aerial image datasets \cite{cao2021visdrone}.
Barmpounakis et al. \cite{barmpounakis2020new} revealed a fundamental mechanism of congestion pattern formation for large-scale networks based on a complete pNEUMA dataset collected by a swarm of drones.
Lin et al. \cite{lin2020vaid} introduced a vehicle detection dataset VAID with the aerial images under different illumination conditions and viewing angles from different places in Taiwan.
Breuer et al. \cite{breuer2020opendd} introduced a large-scale roundabout drone dataset tailored to capturing traffic dynamics in roundabout scenarios. 
Similarly, Krajewski et al. \cite{krajewski2020round} presented the drone-based roundabout drone dataset consisting of the trajectories of road users in Germany.
Zheng et al. \cite{zheng2024citysim} introduced the CitySim dataset with the core objective of facilitating safety-oriented research and applications.

\subsection{Unmanned Aerial Vehicles in Traffic Flow Analysis}

Over the last decade, aerial image processing has been a very popular research topic mainly for the increased data availability \cite{salvo2014urban}. 
Aerial images provide better perspective and can cover a large area in every frame, with the advantage of being mobile, thereby creating a dynamic data collection system \cite{lee2015examining}. 
Thus, in the past few years, UAVs has been an active area of research for different purposes, such as traffic monitoring and traffic management \cite{apeltauer2015automatic}.
Barmpounakis et al. \cite{barmpounakis2016unmanned} demonstrated that, despite few technological obstacles, these systems can be employed for real time traffic monitoring and for the extraction of vehicular trajectories from a video image processing system to collect vehicle traffic data.
Guido et al. \cite{guido2016evaluating} have evaluated the accuracy of vehicle trajectory data at roundabouts obtained via UAVs.
Abdel-Aty et al. \cite{abdel2023advances} discussed the benefits of applying UAVs with computer vision techniques to traffic safety analysis.
Khan et al. \cite{khan2020smart} proposed a traffic monitoring system using the power of UAVs and new emerging technology 5G to monitor, track, and control the speed limit, other traffic violations, and suspicious behavior of moving vehicles on roads and highways.
Zhou et al. \cite{zhou2018computation} proposed a UAV-enabled wireless power supply system for edge computing to address the problem of severe propagation loss of wireless power transmission equipment.
Duan et al. \cite{duan2023joint} established a networked multi-vehilce platoon system utilizing UAV assistance.
Jin et al. \cite{jin2023identifying,jin2025assessing} identified the specific threshold for rear-end conflicts under different weather conditions and investigated the impact of the proximity between signalized and unsignalized intersections within a short-term traffic context based on the CitySim drone dataset.

\section{Methodology} 
\label{sec:2}

This section outlines the methodology of the proposed UAV-based traffic monitoring system, detailing the system architecture, key components, and the processes for vehicle detection, tracking, and data analysis. 
The approach integrates advanced imaging technology, state-of-the-art computer vision algorithms, and robust data analytics to enable real-time traffic monitoring. 
The system is designed to be scalable, interoperable with existing traffic management systems, and resilient under varying environmental conditions.

\subsection{System Architecture}
The proposed UAV-based traffic monitoring system is an modular and integrated platform that combines UAVs-enabled real-time aerial data collection, computer vision-based data processing, and intelligent traffic analytics. 
The overall architecture is shown in Figure \ref{fig1:framework}.
Each of these components is elaborated below.


\subsubsection{Aerial Image Acquisition}
\textbf{(1) Hardware and Flight Dynamics:}
The UAVs are equipped with high-precision inertial measurement units (IMUs) and global positioning system (GPS) modules to follow preplanned flight trajectories that optimize coverage over critical road networks and intersections. 
These trajectories are derived from geospatial analyses to maximize field-of-view while balancing spatial resolution with coverage area. 
The UAVs adjust altitude, orientation, and speed dynamically to adapt to real-time traffic conditions and potential environmental interferences (e.g., wind or rain). 
While the current system operates with preplanned flight paths and offline processing, its modular design and onboard sensing integration lay the foundation for future extensions toward a fully autonomous, real-time aerial traffic monitoring framework capable of in-flight decision-making and adaptive behavior.
\textbf{(2) Imaging Sensors and Data Quality:}
UAVs are equipped with high-resolution imaging sensors, such as 4K or higher cameras, that provide high dynamic range and low-light sensitivity. 
The sensors are integrated with stabilized gimbals to mitigate motion blur, ensuring clear imagery even during high-speed maneuvers. 
The high frame rate (up to 60 fps) is critical for capturing fast-moving vehicles accurately.
\textbf{(3) Communication and Data Management:}
A robust communication system utilizes LTE/5G networks or dedicated RF channels to transmit data from the UAVs to the ground control station. 
The captured video data are compressed using advanced codecs (e.g., H.265/HEVC) to minimize bandwidth requirements while preserving image quality, ensuring near real-time data processing and traffic assessment.

\subsubsection{Computer Vision-Based Processing}
The raw video frames undergo an extensive preprocessing sequence designed to enhance image quality and prepare the data for reliable vehicle detection.
The key preprocessing steps include:
\textbf{(1) Noise Reduction:} Filtering techniques such as Gaussian or median filtering are applied to mitigate sensor noise and environmental artifacts.
\textbf{(2) Grayscale Conversion:} Converting the images to grayscale reduces computational complexity while preserving essential edge and texture information critical for object detection.
\textbf{(3) Adaptive Thresholding:} Foreground segmentation is achieved using methods like Otsu’s thresholding or local adaptive thresholding, which effectively isolate vehicles from the background, enhancing object boundary definition.
\textbf{(4) Contrast Enhancement:} Techniques like histogram equalization or contrast limited adaptive histogram equalization (CLAHE) are employed to accentuate the vehicle features, ensuring robust detection even in challenging lighting conditions.

\subsubsection{Traffic Data Analysis and Visualization}
\textbf{(1) Extraction of Traffic Attributes:}
Following vehicle detection, key traffic metrics are derived from the spatial-temporal data:
\textit{(a) Vehicle Counting and Density Estimation:} Vehicles are counted within defined spatial regions and time intervals to compute traffic density metrics.
\textit{(b) Speed Estimation:} 
Vehicle speeds are estimated by tracking their displacement across consecutive frames. 
To convert pixel-level displacement to real-world distances, we apply a homography transformation calibrated using manually selected ground reference points (e.g., lane widths, crosswalk corners). 
This mapping enables accurate speed estimation in meters per second by projecting image coordinates onto the physical ground plane.
\textit{(c) Trajectory Analysis:} Multi-objective tracking algorithms are used to associate detections across frames, reconstructing trajectories and facilitating analysis of vehicle behavior, lane usage patterns, and intersection dynamics.
In our implementation, we employ frame-to-frame association using Kalman filtering combined with the Hungarian algorithm for data association, effectively tracking vehicles through dense urban scenes. 
This approach supports multi-objective tracking by optimizing both spatial consistency and trajectory smoothness. 
While advanced behavior modeling (e.g., lane-level decision inference or driver intent prediction) is beyond the scope of this work, our trajectory data enables such analyses in future extensions.
\textbf{(2) Statistical and Machine Learning Analytics:}
The extracted parameters are subjected to further analysis using:
\textit{(a) Time-Series Analysis:} Statistical models are employed to identify temporal trends, periodic trends, and anomalies in traffic flow.
\textit{(b) Clustering and Pattern Recognition:} Spatial clustering and pattern recognition algorithms identify congestion hotspots and abnormal traffic incidents, supporting real-time traffic management and informing urban infrastructure planning.
Preliminary implementations of these analytics are conducted through statistical analysis of vehicle distribution profiles and inter-class correlation patterns across multiple observation points. 
These early-stage results illustrate the viability of pattern recognition based on the system's structured outputs and lay the groundwork for integrating more advanced machine learning modules in future developments.

\subsection{Vehicle Detection and Tracking}
Accurate vehicle detection and tracking are fundamental for traffic flow analysis. 
The system employs advanced techniques, such as multi-scale and multi-angle template matching, non-maximum suppression, Kalman filtering, and rotated bounding boxes.

\subsubsection{Multi-Scale and Multi-Angle Template Matching}
Aerial traffic monitoring presents unique challenges due to variations in vehicle size, orientation, and perspective distortion. 
Unlike ground-based images, where vehicles are captured from a consistent viewpoint, UAV imagery introduces scale variations and rotation effects. 
To address these issues, a multi-scale and multi-angle template matching approach is employed, which enhances the robustness of vehicle detection across diverse conditions.
Although deep learning-based object detection frameworks such as YOLO have achieved state-of-the-art accuracy, this study adopts a multi-scale and multi-angle template matching approach due to several practical and system-level considerations. 
UAV-based traffic monitoring systems typically operate under stringent onboard computational constraints, often lacking access to GPUs or dedicated inference accelerators. 
Template matching offers a lightweight and resource-efficient solution suitable for real-time deployment on such platforms. 
Additionally, deep learning models generally require extensive labeled training datasets, which are often unavailable or hard to generalize across diverse aerial perspectives, altitudes, and illumination conditions in dynamic urban environments. 
Furthermore, template matching provides a transparent and interpretable detection mechanism, facilitating system validation and performance diagnostics. 
The proposed framework is modular in nature, allowing for the seamless integration of deep learning-based components in future implementations when computational resources and annotated data become available.

\textbf{(1) Multi-Scale Template Matching:}
Vehicles in aerial imagery appear at varying scales due to altitude, perspective, and camera resolution. 
A fixed-size template is insufficient to detect vehicles at all scales, necessitating a multi-scale approach operating as follows:
\textit{(a) Pyramid Representation:}
Both the input image \textit{I} and the vehicle template \textit{T} are resized to multiple scales using an image pyramid.
This can be achieved using Gaussian or Laplacian pyramids, which systematically downsample the image while preserving critical structural details.
Let \textit{$I_s$} and \textit{$T_s$} denote the versions of \textit{I} and \textit{T} at scale \textit{s}.
This multi-scale representation ensures that features corresponding to vehicles are captured regardless of their apparent size in the scene.
\textit{(b) Cross-Correlation for Detection:}
For each scale \textit{s}, the system computes the normalized cross-correlation (NCC) between \textit{$I_s$} and \textit{$T_s$}.
The NCC function is defined as:
\begin{equation}
\begin{aligned}
&NCC(x,y,s) = \\
&\frac{\sum_{u,v} (I_s(x+u,y+v)-\bar{I}_s(x,y))(T_s(u,v)-\bar{T}_s)}{\sqrt{\sum_{u,v} (I_s(x+u,y+v)-\bar{I}_s(x,y))^2 \sum_{u,v} (T_s(u,v)-\bar{T}_s)^2}}
\end{aligned}
\end{equation}
where, $\bar{I}_s(x,y)$ and $\bar{T}_s$ represent the mean intensity values of the local region in $I_s$ and the entire template $T_s$ respectively.
This normalized measure is robust against variations in illumination and contrast.
\textit{(c) Scale Selection:}
Once the NCC is computed across all scales, the highest correlation value is selected as the optimal match to ensure accurate detection.
To handle multiple vehicles within a single frame, the system computes normalized cross-correlation across all scales and orientations and identifies all local maxima in the correlation response map that exceed a predefined detection threshold. 
These peaks correspond to candidate vehicle locations with strong template similarity. 
To suppress redundant responses and improve localization precision, non-maximum suppression is applied, allowing for accurate detection of multiple vehicles without relying solely on a single global maximum.
This maximum indicates both the best matching location and the most appropriate scale at which the template aligns with a potential vehicle. 
Such a strategy ensures that vehicles are accurately detected irrespective of their size variations in the aerial imagery.

\textbf{(2) Multi-Angle Template Matching}
In aerial imagery, vehicles may be rotated due to the camera’s viewpoint or road curvature. 
Standard template matching, which relies on a fixed-orientation template, struggles in these scenarios and may lead to a significant number of false negatives. 
To address this challenge, the system implements a multi-angle template matching strategy that robustly handles rotated objects:
\textit{(a) Rotated Template Generation:}
Instead of using a single template \textit{T}, a series of rotated templates \textit{$T(\theta)$} are generated, where \textit{$\theta$} spans a predefined range of angles (e.g., \textit{$-\theta_{max}$} to \textit{$+\theta_{max}$}) with a specified angular resolution.
Each rotated template is produced by applying a rotation transformation using a rotation matrix:
\begin{equation}
\left[
\begin{array}{c}
     x' \\
     y' 
\end{array}
\right]
=
\left[
\begin{array}{cc}
     cos\theta & -sin\theta\\
     sin\theta & cos\theta 
\end{array}
\right]
\left[
\begin{array}{c}
     x \\
     y 
\end{array}
\right]
\end{equation}
This process ensures that the orientation variations of vehicles are effectively captured.
\textit{(b) Cross-Correlation across Angles:}
For each rotated template \textit{$T(\theta)$}, the system computes the NCC between \textit{$T(\theta)$} and the input image (or its approprite scaled version, when combined with the multi-scale approach).
The NCC function is applied analogously:
\begin{equation}
\begin{aligned}
 &NCC(x,y,\theta) = \\
 &\frac{\sum_{u,v} (I(x+u,y+v)-\bar{I}(x,y))(T(u,v)-\bar{T})}{\sqrt{\sum_{u,v} (I(x+u,y+v)-\bar{I}(x,y))^2 \sum_{u,v} (T(u,v)-\bar{T})^2}}
\end{aligned}
\end{equation}
By evaluating the response for each angle, the method identifies which rotation of the template best matches a vehicle's orientation in the image.
\textit{(c) Angle Selection and Integration:}
The maximum correlation value across all considered angles is chosen as the best match. 
This selection process not only confirms the presence of a vehicle but also determines its orientation. 
When integrated with the multi-scale matching approach, the system simultaneously determines the optimal scale and orientation, leading to more precise localization and improved detection performance. 
Besides, several optimization strategies, including bilinear interpolation for rotation and parallel processing, are employed to improve efficiency and reduce computation time.

Integrating both multi-scale and multi-angle template matching enables the system to robustly detect vehicles under a wide range of imaging conditions. 
By accurately determining the optimal scale and orientation for each detection, the system enhances the precision of vehicle localization, which is critical for downstream tasks such as tracking, speed estimation, and traffic flow analytics.

\subsubsection{Non-Maximum Suppression for Object Detection}
After detecting vehicles, non-maximum suppression (NMS) is applied to mitigate the issue of duplicate detections and reduces false positives. 
NMS filters out overlapping bounding boxes based on their confidence scores and intersection over union (IoU), which ensures that each vehicle is detected with a single bounding box.
The confidence-driven filtering process is shown as follows:

\textbf{(1) Ranking by Confidence:}
Initially, all candidate bounding boxes are ranked in descending order based on their associated confidence scores, which quantify the likelihood that a given box accurately encapsulates an object.

\textbf{(2) Iterative Suppression:}
The algorithm selects the highest-scoring bounding box as the provisional detection for an object and then computes the IoU between this box and all remaining candidates. 
Any box whose IoU with the selected box exceeds a predefined threshold (commonly in the range of 0.3 to 0.5) is considered redundant and is suppressed. 
This iterative process continues until all boxes have either been selected or suppressed, ensuring that each object is represented by a single, optimal bounding box.

\textbf{(3) Threshold Optimization:}
The selection of the IoU threshold is a critical parameter—if set too low, it may inadvertently discard valid detections, whereas a high threshold might fail to eliminate significant overlaps. 
Our methodology involves an empirical tuning of this threshold through cross-validation, optimizing the balance between precision and recall, particularly in high-density scenarios like urban traffic monitoring.

\textbf{(4) Variants and Enhancements:}
Beyond the standard NMS approach, advanced variations such as Soft-NMS have been explored. 
Unlike traditional NMS that outright removes overlapping boxes, Soft-NMS attenuates the confidence scores of overlapping candidates based on their IoU, allowing a more nuanced retention of potentially valid detections. 
This approach can be particularly beneficial in complex traffic scenes where vehicles are in close proximity, ensuring that subtle but important detections are not prematurely eliminated.

By integrating NMS into our detection pipeline, we achieve a cleaner, more precise set of detections that are essential for subsequent tasks such as vehicle tracking, speed estimation, and trajectory analysis. 
This not only enhances the overall reliability of the system but also reduces computational overhead in downstream processing stages.

\subsubsection{Kalman Filtering for Vehicle Tracking}
The Kalman filter is used to track vehicle movement across frames, especially in scenarios with partial occlusions or rapid vehicle maneuvers. 
By predicting the vehicle’s state and fusing it with new sensor measurements, the Kalman filter improves tracking accuracy and stability.

\textbf{(1) State-Space Modeling:}
The Kalman filter is a recursive Bayesian estimator that operates on a state-space representation of dynamic systems.
In the context of vehicle tracking, the state vector is designed to encapsulate the vehicle's essential motion parameters, typically its position, velocity, and acceleration.
For instance, in a two-dimensional tracking scenario, the state vector can be defined as:
\begin{equation}
\textbf{x}_k =
\left[
\begin{array}{c}
     x_k \\
     y_k \\
     \dot{x}_k \\
     \dot{y}_k
\end{array}
\right]
\end{equation}
where, \textit{$x_k$} and \textit{$y_k$} denote the spatial coordinates of the vehicle at time step \textit{$k$}, and \textit{$\dot{x}_k$} and \textit{$\dot{y}_k$} represent the corresponding velocity components.

Under the assumption of a constant velocity model, the state evolution is modeled as:
\begin{equation}
\textbf{x}_k = \textbf{F}\textbf{x}_{k-1} + \textbf{w}_k
\end{equation}
where:
\begin{itemize}
    \item $\textbf{F}$ is the state transition matrix that encapsulates the dynamics of the system. For a constant velocity model, $\textbf{F}$ is typically structured as: 
    \begin{equation}
\textbf{F} =
\left[
\begin{array}{cccc}
     1 & 0 & \Delta t & 0 \\
     0 & 1 & 0 & \Delta t \\
     0 & 0 & 1 & 0 \\
     0 & 0 & 0 & 1
\end{array}
\right]
\end{equation}
with \textit{$\Delta t$} representing the the time interval between successive measurements. This formulation predicts that the new position is the sum of the previous position and the product of velocity with the elapsed time.
    \item $\textbf{w}_k$ is the process noise vector, which models the uncertainty inherent in the system dynamics. It is assumed to be a zero-mean Gaussian noise: 
    \begin{equation}
    \textbf{w}_k \Rightarrow \mathcal{N}(0,\textbf{Q})
    \end{equation}
with $\textbf{Q}$ being the process noise covariance matrix. The selection of $\textbf{Q}$ is critical, as it determines how much uncertainty is attributed to the model's predictions relative to the measurements.
\end{itemize}

This state-space formulation forms the basis for robust vehicle tracking, allowing the Kalman filter to propagate state estimates even in the presence of missing or noisy observations.

\textbf{(2) Prediction and Update Cycle:}
The Kalman filter operates recursively through two fundamental steps, i.e. prediction and update, which together enable continuous refinement of the state estimate as new measurements are acquired.
\textit{(a) Prediction Step:}
In the prediction phase, the filter uses the state transition model to compute a prior estimate of the state at time \textit{$k$} based on the most recent state estimate at time $\textit{k}$-1.
This step can be mathematically expressed as:
\begin{equation}
\hat{\textbf{x}}_{k|k-1} = \textbf{F}\hat{\textbf{x}}_{k-1|k-1}
\end{equation}
where, $\hat{\textbf{x}}_{k|k-1}$ denotes the predicted state given all measurements up to time $\textit{k}$-1.
Simultaneously, the error covariance associated with the state estimate is predicted to account for the propagation of uncertainty through the dynamic model:
\begin{equation}
\textbf{P}_{k|k-1} = \textbf{F}\textbf{P}_{k-1|k-1}\textbf{F}^T + \textbf{Q}
\end{equation}
where, $\textbf{P}_{k-1|k-1}$ is the error covariance of the state estimate at time $\textit{k}$-1 and $\textbf{Q}$ captures the uncertainty due to process noise. 
This predicted covariance matrix $\textbf{P}_{k|k-1}$ quantifies the expected error in the predicted state, effectively bridging the gap between frames, especially during periods where direct detection might be unreliable due to occlusion or abrupt maneuvers.
\textit{(b) Update Step:}
Once a new measurement $\textbf{z}_k$ becomes available, typically extracted from an object detection pipeline, the filter performs an update to correct the predicted state.
The measurement model is given by:
\begin{equation}
\textbf{z}_k = \textbf{H}\textbf{x}_{k} + \textbf{v}_k
\end{equation}
where:
\begin{itemize}
    \item $\textbf{H}$ is the measurement matrix that maps the state vector into the observation space. For example, if the measurement directly observes the position, then $\textbf{H}$ is defined as: 
    \begin{equation}
\textbf{H} =
\left[
\begin{array}{cccc}
     1 & 0 & 0 & 0 \\
     0 & 1 & 0 & 0
\end{array}
\right]
\end{equation}
    \item $\textbf{v}_k$ is the measurement noise vector, also assumed to be a zero-mean Gaussian noise: 
    \begin{equation}
    \textbf{v}_k \Rightarrow \mathcal{N}(0,\textbf{R})
    \end{equation}
with $\textbf{R}$ being the measurement noise covariance matrix.
\end{itemize}
The first step in the update process is to compute the Kalman gain $\textbf{K}_k$, which quantifies the relative weight given to the prediction versus the new measurement:
\begin{equation}
\textbf{K}_k = \textbf{P}_{k|k-1}\textbf{H}^T(\textbf{H}\textbf{P}_{k|k-1}\textbf{H}^T + \textbf{R})^{-1}
\end{equation}
The Kalman gain acts as an adaptive blending factor: if the predicted covariance is high relative to the measurement noise, $\textbf{K}_k$ will be larger, meaning that the new measurement significantly influences the state update.
Conversely, if the measurement noise is high, the filter will rely more heavily on the prediction.
Using the computed Kalman gain, the predicted state is then updated to incorporate the measurement:
\begin{equation}
\hat{\textbf{x}}_{k|k} = \hat{\textbf{x}}_{k|k-1} + \textbf{K}_k(\textbf{z}_k - \textbf{H}\hat{\textbf{x}}_{k|k-1})
\end{equation}
where, the term $\textbf{z}_k - \textbf{H}\hat{\textbf{x}}_{k|k-1}$ is known as the measurement residual, representing the discrepancy between the observed measurement and the predicted measurement.
Finally, the error covariance is updated to reflect the improved state estimate:
\begin{equation}
\textbf{P}_{k|k} = (\textbf{I} - \textbf{K}_k\textbf{H})\textbf{P}_{k|k-1}
\end{equation}
This updated covariance $\textbf{P}_{k|k}$ is typically reduced relative to the predicted covariance, indicating the uncertainty has been partially corrected through the measurement update.

By leveraging the state-space representation and recursively applying the prediction and update cycles, the Kalman filter provides a powerful framework for vehicle tracking.
The state-space model captures the essential dynamics of vehicle motion, while the recursive cycle of prediction and update continuously refines the state estimate.
This approach effectively manages uncertainty from both the process dynamics and noisy measurements, ensuring robust and accurate tracking even in challenging envirnments with occlusions or rapid movements.

\textbf{(3) Handling Occlusions and Rapid Movements:}
In complex traffic scenarios, vehicles may experience temporary occlusions or abrupt accelerations that cause intermittent or imprecise detections. 
During these periods, the Kalman filter’s prediction step plays a pivotal role by propagating the vehicle’s estimated state forward, thus maintaining a continuous track despite missing or noisy observations. 
This capability minimizes the likelihood of track fragmentation and enhances the reliability of vehicle trajectory reconstruction.

\textbf{(4) Parameter Tuning and Adaptation:}
The effectiveness of the Kalman filter hinges on the accurate tuning of the process noise covariance $\textbf{Q}$ and the measurement noise covariance $\textbf{R}$. 
These parameters are empirically optimized through cross-validation using real-world traffic data, ensuring that the filter remains adaptive to varying dynamic conditions, such as dense traffic clusters or high-speed scenarios. 
Fine-tuning these parameters directly impacts the responsiveness of the filter in balancing trust between model predictions and new observations.

In summary, the application of the Kalman filter in our vehicle tracking module significantly improves the system’s resilience to measurement noise and transient occlusions. 
By continuously predicting and updating vehicle states, the filter ensures smoother and more consistent trajectories, which are critical for downstream analyses such as speed estimation, trajectory prediction, and comprehensive traffic flow analytics.

\subsubsection{Rotated Bounding Boxes for Vehicle Orientation}
To better capture vehicles’ orientation, especially on curving roads, the system employs rotated bounding boxes instead of conventional axis-aligned boxes. 

\begin{figure*}[ht]
    \centering
    \includegraphics[width=0.8\textwidth]{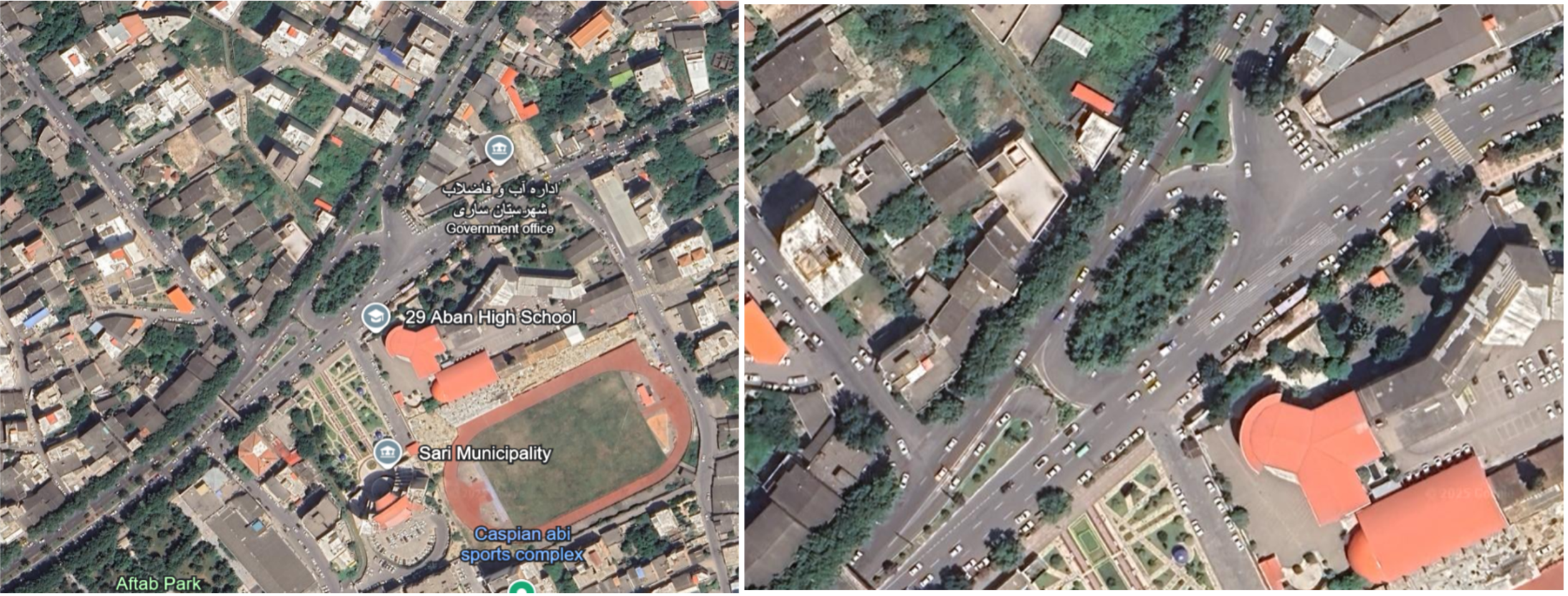}
    \caption{Satellite imagery of the selected intersection in central Sari: 
    \textbf{(1) Wide-area view} showing the broader urban context of the intersection, located at the convergence of Azadi Boulevard, Ferdowsi Street, Amir Mazandarani Street, and 20-Metri Gol Afshan.
    \textbf{(2) Close-up satellite image} illustrating the geometric layout of the roundabout and its surrounding roads used for UAV-based traffic analysis.}
    \label{fig3:Satellite}
\end{figure*}

\textbf{(1) Enhanced Representation:}
Rotated bounding boxes are generally parameterized by a five-dimensional vector.
This enriched representation enables the bounding box to conform more closely to the actual shape and orientation of a vehicle, reducing the inclusion of background clutter and improving the overall detection precision.

\textbf{(2) Alignment with Vehicle Orientation:}
Vehicles, especially on curving roads, are often oriented at angles that deviate significantly from the image axes. 
Axis-aligned boxes may therefore capture extraneous background information or even overlap adjacent vehicles, leading to lower IoU scores during evaluation. 
By employing rotated bounding boxes, the detection framework aligns more naturally with the vehicles’ orientations, enhancing both the localization accuracy and the quality of the extracted features for downstream tracking and classification tasks.

\textbf{(3) Vehicle Classification Method:}
Following detection and orientation alignment, vehicles are classified into five categories: motorcycle, car, pickup, bus, and taxi based on a set of visual features extracted from the localized bounding boxes. 
The classification method combines template-based shape matching, aspect ratio analysis, and relative size thresholds, further refined by color cues when available.
Larger bounding boxes that exceed predefined area thresholds are assigned to heavy vehicle classes such as bus and pickup, while smaller objects are differentiated based on their aspect ratio and contour shape. 
This rule-based heuristic approach avoids the need for large-scale labeled datasets, provides transparent and interpretable classification results, and ensures computational efficiency suitable for real-time UAV applications.
The resulting class labels are used to generate class-wise traffic statistics, inform behavioral analysis, and support higher-level decision tasks in subsequent processing stages. 
Future work may explore the integration of learned classifiers to improve scalability and generalization.

\textbf{(4) Improved Tracking Accuracy:}
Accurate tracking is highly dependent on precise localization from frame to frame. 
On curved roads, vehicles continuously change orientation, which can result in mismatches when using standard axis-aligned boxes. 
Rotated bounding boxes maintain a consistent representation of each vehicle, thereby reducing spatial errors in successive frames, facilitating smoother trajectory estimation, and enhancing the performance of subsequent tracking algorithms such as Kalman filters by providing more reliable and consistent input data.

\textbf{(5) Adaptation in Post-Processing:}
The integration of rotated bounding boxes necessitates modifications in the post-processing pipeline:
\textit{(a) Rotated IoU Computation:}
Standard IoU calculations are extended to consider the rotation angle, often referred to as rotated IoU. 
This metric more accurately reflects the overlap between rotated boxes, ensuring that detection refinement processes, such as NMS, effectively suppress redundant or false-positive detections.
\textit{(b) Anchor Generation and Regression:}
During network training, rotated anchors covering a range of angles are generated. 
The loss function is adapted to penalize discrepancies not only in the bounding box location and scale but also in the angular orientation, leading to improved convergence and detection accuracy.

\textbf{(6) Robustness to Perspective Distortion:}
In aerial imaging scenarios, perspective distortion is a common challenge that can cause vehicles to appear rotated relative to the camera’s viewpoint. 
Rotated bounding boxes naturally account for such distortions, providing consistent and accurate vehicle delineation regardless of the angle at which the vehicle is observed. 
This is particularly important in UAV-based traffic monitoring, where varying flight angles and altitudes can introduce significant variations in the captured imagery.

By incorporating rotated bounding boxes into our detection and tracking pipeline, the system achieves a higher degree of precision in vehicle localization and tracking. 
This improvement is especially pronounced in environments characterized by curving roads and dynamic vehicle orientations, ultimately leading to more reliable traffic analysis and enhanced overall system performance.

\subsection{Traffic Analysis}
To quantify traffic dynamics, the system extracts and analyzes key metrics:

\subsubsection{Speed Estimation}
Vehicle speed is computed using a frame-by-frame displacement method, considering the UAV's altitude and camera frame rate. 
A homography transformation is applied to map pixel distances to real-world distances.
While the UAV is capable of dynamic repositioning, for the speed estimation experiments presented in this study, it is operated in a stationary hover mode. 
This operational choice ensures consistent perspective and altitude during video capture, allowing a single homography calibration to be applied without the need for continuous re-estimation.
In scenarios involving camera drift or UAV motion, the current system constrains speed estimation to short, quasi-static intervals where motion is negligible. 
For future work, we plan to implement dynamic homography updates using visual landmarks and road geometry to maintain reliable pixel-to-world mapping during continuous UAV movement.

\subsubsection{Traffic Congestion}
It is acknowledged that the current density-based clustering approach (e.g., DBSCAN) may also identify temporary queues in front of signalized intersections as congestion zones due to high vehicle concentration. 
To address this, the system incorporates a temporal persistence filter, in which only clusters that remain stable beyond a predefined time threshold are flagged as congestion. 
This effectively reduces false positives caused by transient queuing behavior.
In future work, the integration of contextual information, such as signal phase timing and temporal movement trends, will be considered to further distinguish between normal queuing and irregular congestion caused by downstream blockages, incidents, or bottlenecks.

\subsubsection{Traffic Flow}
By tracking vehicle trajectories, the system generates detailed flow maps that illustrate traffic movement patterns across time and space. 
These maps assist traffic authorities in identifying bottlenecks by detecting areas of speed reduction or congestion formation. 
Additionally, the system is capable of recognizing irregular driving behaviors such as sudden stops, abrupt lane changes, or erratic movements that may indicate accidents, traffic violations, or road hazards. 
This information supports real-time incident detection and facilitates more responsive traffic management.

\subsubsection{Violation Detection}
The system automatically detects speed violations and lane-changing irregularities by comparing vehicle behavior against predefined traffic regulations.

\subsubsection{Traffic Demand}
Predicting traffic demand using machine learning remains a valuable direction for future work.
A random forest regression model trained on historical traffic data such as time-of-day, weather conditions, and road layout has the potential to anticipate congestion trends and support proactive traffic management.
However, the development and validation of such predictive models require extensive data preprocessing, parameter tuning, and evaluation against ground truth measurements components that are beyond the current scope. 
As such, traffic demand forecasting is identified as a promising extension for future system iterations aimed at enhancing analytical depth and predictive capability.

\begin{figure*}[ht]
\centering
\includegraphics[width=0.8\linewidth]{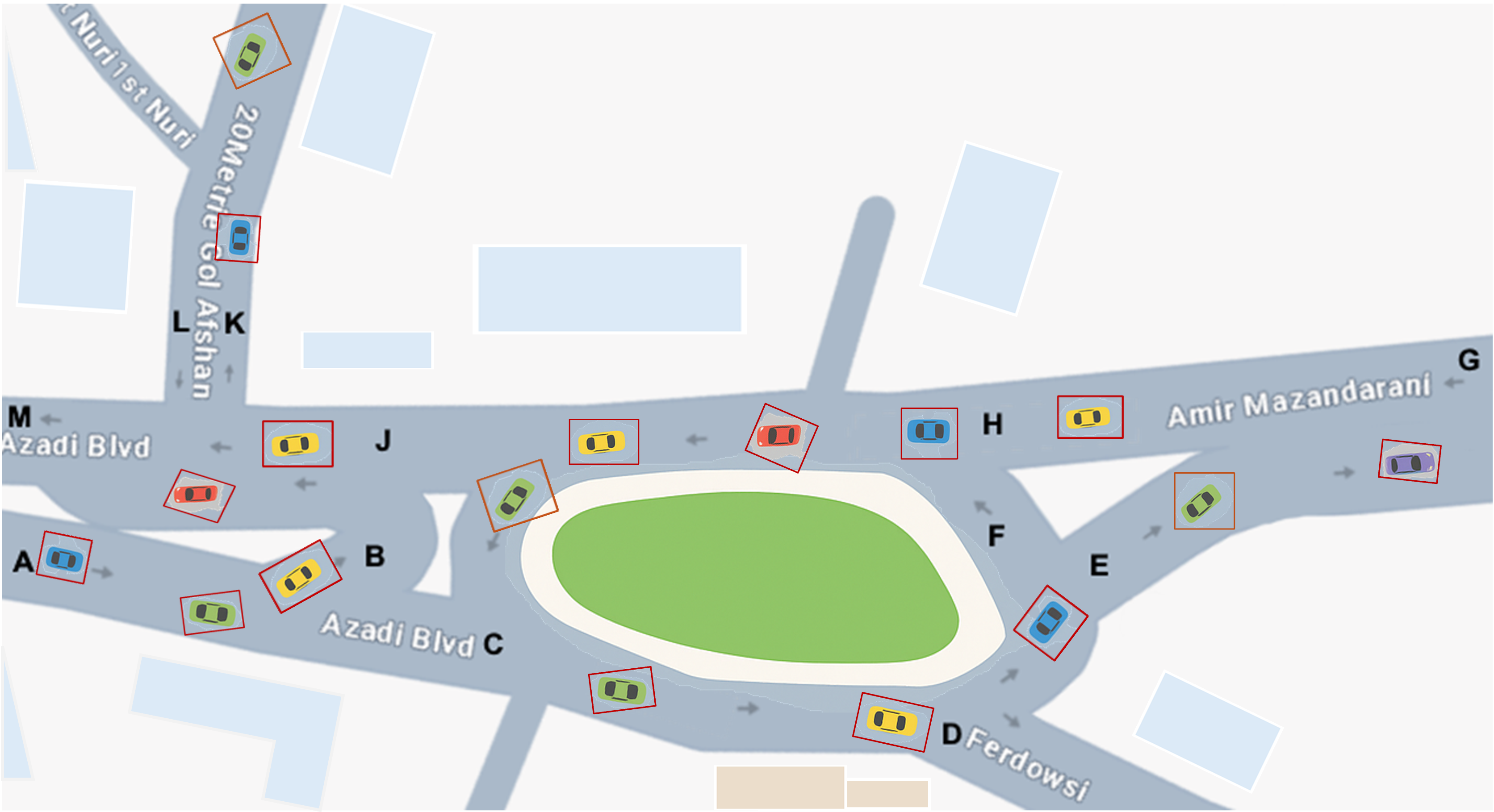}
\caption{Layout of the monitored urban intersection with labeled entry and exit points}
\label{fig8:Layout}
\end{figure*}

\section{Experiments} 
\label{sec:3}

\subsection{Case Study}

\subsubsection{UAV Specifications and Deployment Conditions}

As shown in Figure~\ref{fig2:DJIUAV}, a DJI Air 2S drone was used for data collection in this study. 
The drone features a 1-inch CMOS sensor camera capable of recording 5.4K video at 60 frames per second with a resolution of 3840×2160 pixels, and still images up to 20 MP. 
Its stabilized 3-axis gimbal enables consistent top-down imagery, minimizing motion blur and distortion during flight. 
The drone was operated at a fixed altitude of approximately 200–220 meters  above ground level to ensure full coverage of the intersection with minimal occlusion.

With a maximum flight time of 31 minutes and GPS/GLONASS satellite positioning, the UAV provided stable hovering and smooth flight control. 
All video data was streamed in real time to a connected mobile device and stored locally on a microSD card for post-processing. 
The drone was flown in nadir orientation under clear weather conditions (light wind < 20 km/h), enabling reliable vehicle detection and tracking.

\subsubsection{Study Area Description and Data Collection Conditions}

To assess the effectiveness of the proposed UAV-based traffic monitoring framework in a real-world context, a field experiment was conducted in Sari, the capital of Mazandaran Province in northern Iran (Figure~\ref{fig3:Satellite}). 
With a population exceeding 360,000 and a high influx of vehicles during weekends and holidays due to its administrative and touristic role, Sari frequently experiences heavy traffic volumes—especially in its central business district (CBD).

The selected intersection is located at the convergence of Azadi Boulevard, Ferdowsi Street, Amir Mazandarani Street, and 20-Metri Gol Afshan, with a large central roundabout facilitating traffic circulation. 
This location was strategically chosen due to its critical role in the city's traffic network. 
It lies adjacent to the municipal complex, connects key institutional and commercial zones, and provides the shortest and most direct links between the city center and major regional exits. 
The area is also surrounded by schools, administrative offices, and a stadium, contributing to peak-hour traffic congestion during both routine and event-related periods.

UAV flights were conducted on a weekday during the early evening peak (14:30 to 16:30) under favorable weather and light wind conditions. 
A DJI Air 2S drone was flown at a fixed altitude of 200–220 meters, recording nadir-view 4K video footage. 
The intersection's geometric complexity and strategic importance provided a robust testbed for validating the system’s performance in vehicle detection, tracking, and traffic behavior analysis.

\subsection{Experimental Results}

\subsubsection{Vehicle Detection and Tracking Using Template Matching}
We employed multi-scale template matching, adaptive thresholding, and Kalman filtering to achieve real-time vehicle detection and tracking under various conditions. 
While the current system processes video footage offline, it achieves high frame processing rates that allow us to analyze traffic events efficiently. 
Real-time implementation is not yet deployed; however, the system is designed with modularity in mind to allow future integration with live video streams using HDMI capture or DJI SDK-based methods.

The UAV operated at a stable flight altitude during the entire data collection process. 
This consistent altitude helped maintain uniform image resolution and vehicle size across the video frames, contributing to reliable detection and tracking. 
The surveillance was conducted over a typical urban area that included intersections, U-turn zones, and roadside parking spaces.

\begin{figure*}[ht]
\centering
\includegraphics[width=0.8\linewidth]{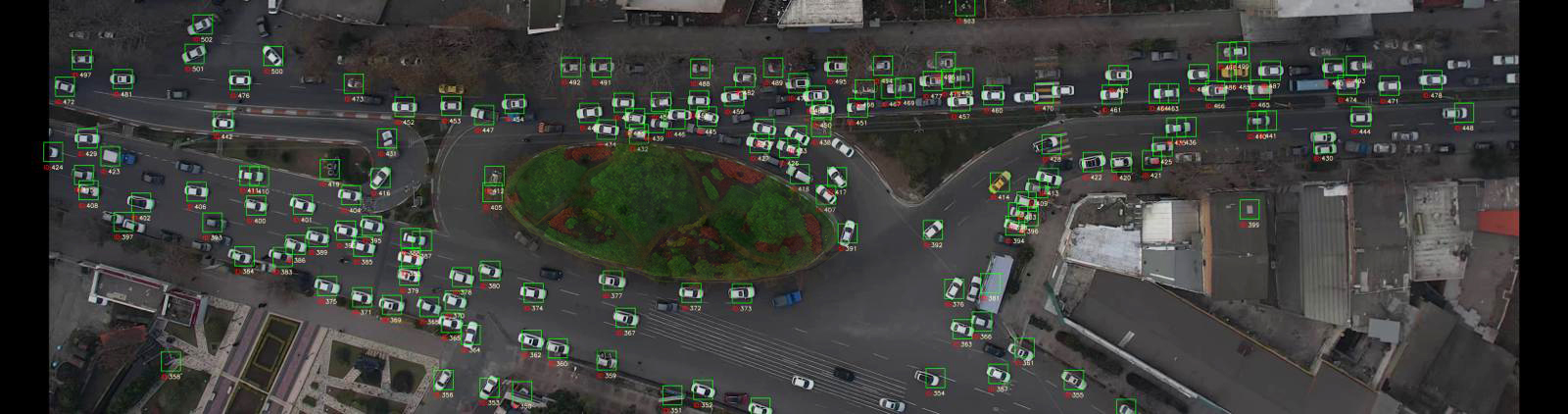}
\caption{UAV-based vehicle detection and tracking in a real urban environment}
\label{fig4:UAVbased}
\end{figure*}

\begin{table}[ht]
\centering
\caption{Detection and tracking performance}
\label{tab:detection}
\begin{tabular}{lc}
\toprule
\textbf{Metrics} & \textbf{Value (\%)} \\
\midrule
Precision & 91.8 \\
Recall & 89.2 \\
F1 Score & 90.5 \\
False Positives & 3.5 \\
False Negatives & 7.5 \\
Overall Detection Accuracy & 90.8 \\
MOTA & 92.1 \\
MOTP & 93.7 \\
\bottomrule
\end{tabular}
\end{table}

Detection and Tracking performance are presented in Table~\ref{tab:detection}. 
The results were tested on 100 labeled frames with manual verification.
3.5 \% of detections were incorrect, where the system mistakenly detected a vehicle where there was none (e.g., shadow or road marking classified as a vehicle).
7.5 \% of actual vehicles were not detected, typically due to occlusion, lighting conditions, or low contrast.
The overall detection accuracy represents the percentage of correctly detected vehicles out of all vehicles present in the test frames.
Tracking accuracy is highly competitive with minimal ID switches.
Direct comparisons with other vehicle detection systems were not included in this study due to variations in dataset conditions, UAV flight parameters, and algorithmic approaches. 
However, we recognize the value of such benchmarks and plan to incorporate them in future versions of the work for broader context and performance positioning.

Figure~\ref{fig4:UAVbased} shows the UAV-based vehicle detection and tracking in a real urban environment.
Bounding boxes illustrate the output of multi-scale template matching and Kalman filtering in dense traffic conditions.
In the final implementation, fixed-size square bounding boxes aligned with the image axes were used instead of rotated ones. 
This approach was selected to simplify the detection pipeline and maintain efficiency across different traffic scenes.

\subsubsection{Speed Estimation}

To estimate vehicle speed in real-world conditions, we defined key reference points (A to B and G to H) within the camera's field of view, as shown in Figure~\ref{fig8:Layout}. 
The real-world distances between these reference points were calculated using a calibrated homograph mapping technique that translates image coordinates into geographic measurements. 
Each vehicle's motion was tracked frame-by-frame to determine its displacement over time. 
By combining the measured displacement in pixels with the known frame rate (FPS) of the video footage, we accurately converted pixel-level movement into real-world speed values. 
The A–B and G–H segments were selected to represent two distinct traffic environments: a high-activity zone near a U-turn and a congested intersection area. 
These paths provided diverse conditions for evaluating the robustness of the speed estimation method under varying traffic flow patterns and vehicle behaviors.

\textbf{(1) A to B Path (Three Lanes, U-Turn Zone): Measured speeds: 12.7 km/h, 13.5 km/h, 12.0 km/h.} 
Speed fluctuations occurred due to lane changes in preparation for a U-turn at point B.
Vehicles slowed down before executing the U-turn, resulting in lower speed readings.
The middle lane showed the highest speed consistency, while outer lanes exhibited greater variation.  

\textbf{(2) G to H Path (Two Lanes, Intersection with Turning Traffic): Measured speeds: 5.31 km/h, 6.24 km/h.} 
Lower average speeds were recorded due to traffic congestion caused by turning vehicles at a nearby square intersection.
Vehicles in the left lane slowed down significantly due to merging or stopping before turns.
Right lane traffic moved slightly faster but still faced periodic slowdowns.

For future work, we plan to visualize vehicle speed profiles using color gradient overlays on the road layout. 
This approach will provide an intuitive representation of speed variations across different lanes and segments, helping to highlight congestion zones and traffic flow behavior.

\subsubsection{Vehicle Type Classification}
Despite the UAVs' relatively high operating altitude (200 meters), the system maintained reliable detection and tracking accuracy, which demonstrates its effectiveness in handling long-range aerial footage.
Table~\ref{tab:classification} presents the classification accuracy. 

\begin{table}[ht]
\centering
\caption{Classification accuracy}
\label{tab:classification}
\begin{tabular}{lccc}
\toprule
\textbf{Vehicle Type} & \textbf{Precision (\%)} & \textbf{Recall (\%)} & \textbf{F1 Score (\%)} \\
\midrule
Motorcycle & 86.7 & 83.9 & 85.3 \\
Taxi & 90.1 & 87.4 & 88.7 \\
Private Car & 94.5 & 92.1 & 93.3 \\
Pickup & 85.9 & 82.7 & 84.2 \\
Bus & 97.2 & 95.6 & 96.4 \\
\bottomrule
\end{tabular}
\end{table}

Bus classification is nearly perfect due to its distinct shape and size.
The overall classification accuracy is 90.3 \%.
The majority of misclassifications occur between Pickup and Private Car.
Occlusion issues in dense traffic reduce recall.
Misclassifications between Pickup and Private Car remain challenging.
High-speed vehicles ($>$100 km/h) show slight tracking drift. 

\subsubsection{Illegal Driving Behavior Detection}

To evaluate compliance with traffic regulations, the proposed system integrates multi-scale template matching, Kalman-based object tracking, and geofencing-based spatial constraints. 
By analyzing vehicle trajectories and identifying abnormal speed patterns, the system can automatically detect common traffic violations. 
These include sudden or unsafe lane changes near U-turn zones, illegal double parking in side streets and near intersections, and vehicle obstruction in designated crosswalk areas.
The system can also be extended to detect speeding violations by comparing estimated vehicle speeds with predefined limits for each segment. 
This would require incorporating road-specific speed data and may enhance the range of detectable traffic infractions.

\textbf{(1) Illegal Lane Change Near U-Turn (Point B):} 
Observations revealed that vehicles are required to switch lanes at least 100 meters before reaching the U-turn at Point B. 
However, our system detected multiple instances of vehicles changing lanes less than 30 meters before the U-turn, which caused sudden slowdowns and increased the risk of collisions. 
To detect such violations, we applied a lane boundary detection algorithm based on image processing and perspective transformation to accurately define lane positions. 
Vehicles were then tracked using Kalman filtering, and trajectory deviations occurring within 100 meters of the U-turn were flagged as violations. 
The violation rate observed in this segment was 16.3 \% of all vehicles passing through the monitored zone. 
This behavior contributes to traffic congestion, particularly when vehicles abruptly cut into the U-turn lane. 
Additionally, detection becomes more challenging when multiple vehicles cluster near the U-turn simultaneously, leading to occlusion-related issues during tracking.

\textbf{(2) Illegal Double Parking on Side Streets and Near Square Intersection:} 
Observations revealed that double parking was frequently detected in two specific areas: (a) side streets near commercial zones within the CBD, and (b) the square intersection—particularly near points A to C and G \& H—where drivers often parked briefly to drop off passengers. 
This behavior blocks traffic flow and creates bottlenecks in high-density areas. 
To detect such violations, we employed a geofencing method in which predefined no-parking zones were established within the surveillance area. 
Any stationary vehicle detected within a restricted area for more than 10 seconds was flagged. 
Motion analysis was also used, where vehicles stopping briefly (less than 10 seconds) were ignored to account for temporary stops such as pedestrian crossings.
Traffic delays were observed to increase by 11–18 \% during peak hours due to double parking, underscoring its significant impact on traffic efficiency. 
To improve robustness against small positional fluctuations, vehicles were considered stationary if their estimated speed remained consistently below a low threshold value (e.g., near-zero speed) for more than 10 seconds within a restricted area. 
This approach ensures reliable double-parking detection while allowing brief temporary stops to be ignored.
While the current system focuses on motor vehicles, future extensions will include the detection of pedestrians and cyclists to support urban safety analysis, particularly in zones with potential high-speed vehicle interaction. 
This enhancement will allow the system to better evaluate the risk to vulnerable road users.
Although we observed a clear temporal and spatial association between certain traffic violations and congestion (e.g., delays following double parking incidents), we recognize that delays can also result from other factors such as signal timing, pedestrian flow, or random fluctuations. 
Future work will incorporate multivariate traffic analysis to better isolate the specific causes of delays.

\textbf{(3) Illegal Parking in Crosswalk Areas:} 
Observations revealed that vehicles were frequently detected blocking pedestrian crosswalks, including instances involving taxi stops where taxis were stopping illegally. 
To detect these violations, we performed crosswalk region mapping by defining crosswalk areas using homography projection and comparing vehicle bounding boxes against the predefined regions. 
Violation detection was then carried out by flagging vehicles that remained in the crosswalk area for more than 10 seconds. 
The violation rate analysis showed that private cars and taxis were the most frequent offenders, accounting for 42.7 \% of the total recorded violations. 
This behavior posed a direct pedestrian safety risk, with 4.8 \% of the violations forcing pedestrians to step onto the road outside the designated crossing area. 
Detection challenges were also noted, particularly under heavy congestion, which occasionally obstructed clear visibility of the crosswalk zones and highlighted the need for improved occlusion handling mechanisms. 
While the camera provides a top-down view, occlusion occurred when multiple vehicles accumulated near the crosswalk, temporarily blocking parts of the predefined crossing area. 
This made it difficult to detect new violations accurately during congested frames. 
All crosswalk areas were manually defined via homography mapping based on static scene geometry. 
Future work will focus on detecting pedestrians and analyzing vehicle–pedestrian interaction to assess compliance with yield rules at zebra crossings.

To enhance the system's capability in detecting traffic violations, several future improvements are proposed. 
One key enhancement involves integrating deep learning models, such as YOLO, to improve vehicle classification and effectively handle occlusion scenarios. 
Additionally, deploying a real-time alert system is planned to notify law enforcement agencies promptly about recurring violations, thereby facilitating timely interventions. 
Another significant advancement includes combining LiDAR and vision-based tracking technologies to enhance detection accuracy, especially in high-traffic environments.
In summary, the current system successfully detects lane-change violations, double parking, and crosswalk obstructions. 
Data-driven analyses have provided insights into peak violation times and the types of vehicles most frequently involved in infractions. 
Future developments will focus on implementing deep learning-based tracking mechanisms and real-time alert systems to further improve the system's effectiveness in traffic law enforcement.
In addition to classification and violation detection, the system can be extended to perform origin-destination (OD) analysis and traffic flow modeling. 
Since vehicle trajectories are already tracked, mapping their entry and exit points can yield valuable insights for urban planning and congestion mitigation.

\begin{figure*}[ht]
\centering
\includegraphics[width=0.8\linewidth]{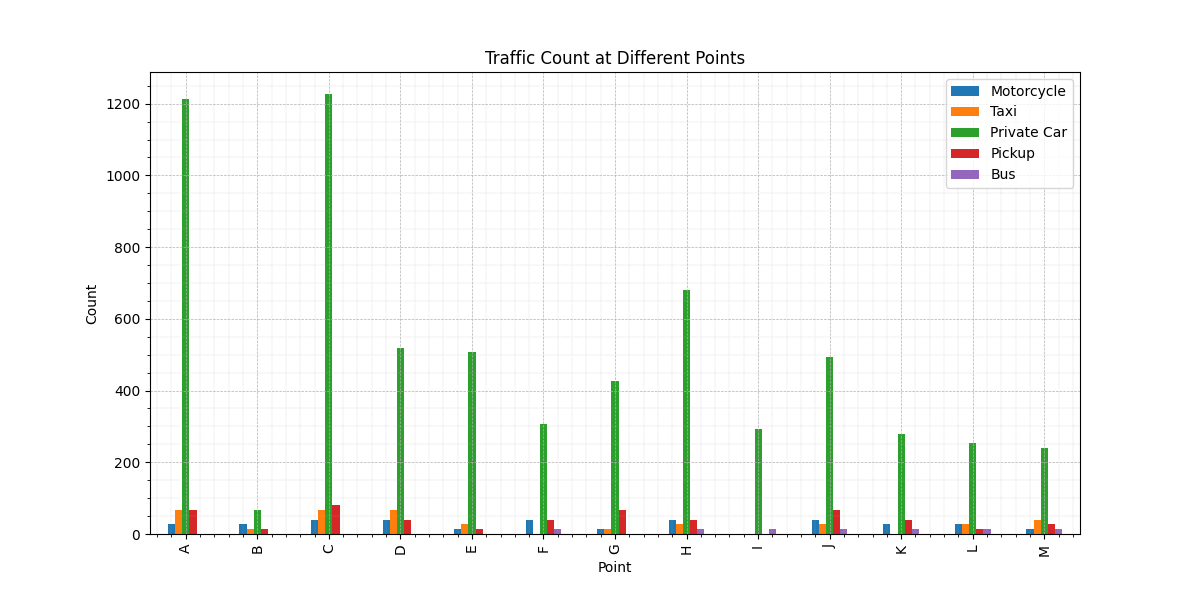}
\caption{Traffic count distribution by vehicle type across monitored points: Five vehicle categories are represented: motorcycles (blue), taxis (orange), private cars (green), pickups (red), and buses (purple).}
\label{fig5:Traffic}
\end{figure*}

\subsubsection{Traffic Count Analysis at Different Points}

The analysis of vehicle counts across various observation points reveals several key trends. 
The traffic counts presented in Figure~\ref{fig5:Traffic} were collected over a continuous observation period of approximately 20 minutes. 
Vehicle counts reflect the total number of vehicles observed during the full duration of each recording session. 
While standard normalization to fixed time windows (e.g., 15 or 60 minutes) is not applied here, future work will incorporate such normalization for enhanced comparability across observation points and time intervals.

\textbf{(1) Dominance of Private Cars:} 
At all locations, private cars (green bars) are the most frequently detected vehicles, with significantly higher counts compared to other vehicle types. 
Points A and C show peak private car counts (1200 vehicles each), indicating high traffic density at these locations. 
This suggests that these locations may be major entry/exit points, high-traffic roads, or key intersections.

\textbf{(2) Traffic Distribution Variation Across Points:}
Some points, such as D, E, H, and J, show moderate vehicle flow, with private car counts ranging between 400–700. 
Other locations, like B, F, and M, have noticeably lower traffic volumes, indicating either restricted vehicle flow or alternative road usage patterns.
The observed differences in traffic volumes may reflect a functional distinction between through-traffic routes and locally-oriented streets. 
While this study does not explicitly classify traffic flows by purpose, future work will explore origin–destination analysis to further investigate this hypothesis and better understand spatial usage patterns.

\textbf{(3) Pickup and Taxi Usage Patterns:}
Taxis (orange bars) and pickups (red bars) are present at all locations but in smaller numbers compared to private cars. 
Pickups are more frequent in commercial areas or delivery zones, while taxis are more concentrated near high-traffic regions (A, C, H, J). 
Misclassification issues between pickups and private cars could be a potential challenge in our classification model.

\textbf{(4) Bus Traffic Trends:}
Buses (purple bars) are relatively rare across all points. 
This is expected since public transportation typically operates in designated lanes and specific routes. 
However, the presence of buses at multiple points suggests that public transport infrastructure is actively used in the study area. 

\textbf{(5) Potential Traffic Congestion Points:}  
The extremely high private car counts at Points A and C may reflect high traffic demand. 
However, this does not necessarily imply congestion, as efficient traffic management may allow for smooth flow. 
These areas could still warrant further investigation to assess flow performance, capacity utilization, and potential needs for optimization.

\subsubsection{Gradient Heatmap Analysis of Traffic Counts}

\begin{figure}[ht]
\centering
\includegraphics[width=\linewidth]{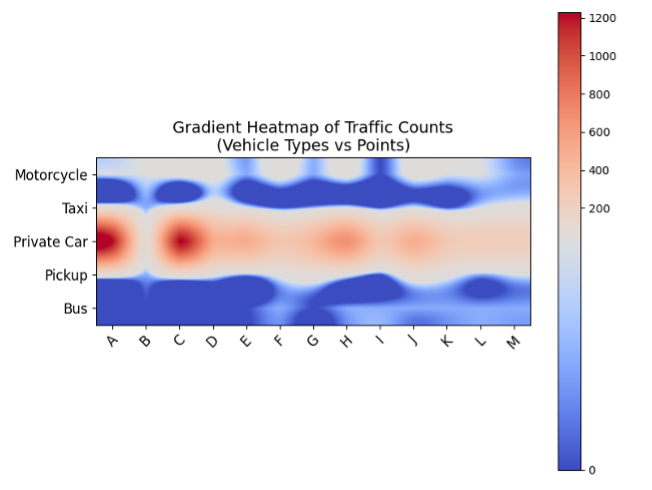}
\caption{Gradient heatmap of vehicle count intensity by type and location}
\label{fig6:Gradient}
\end{figure}

The heatmap in Figure~\ref{fig6:Gradient} does not represent geographic space, but rather shows the distribution of vehicle types across observation points. 
Each cell reflects count intensity for a specific vehicle type at a specific location. 
A redesigned grid-based layout with clarified color scaling is planned for future versions. 
The heatmap currently reflects total vehicle counts, not time-normalized intensities. 
Future work will apply temporal normalization (e.g., per 15-minute or 1-hour windows) to enable better comparability between locations and time periods.

\textbf{(1) Private Cars Dominate at Most Locations:}
The most intense red regions correspond to private cars, confirming their dominant presence in the dataset.
The highest concentrations appear at Points A, C, and H, indicating heavy traffic flow at these locations.

\textbf{(2) Motorcycle and Bus Activities are Sparse:}
The motorcycle row is mostly blue with a few scattered lighter regions, indicating sporadic motorcycle presence.
The low number of bus detections, as indicated by the blue tones in the bottom row of the heatmap, reflects the relatively rare presence of buses in the observed area rather than a detection issue.
This suggests that public transport usage in the study area is relatively low or follows dedicated lanes not captured by our system. 

\textbf{(3) Pickups Show Mid-Level Distribution:}
The pickup category has some moderate intensity regions at a few points (G, J), possibly indicating commercial vehicle movement.
This may be linked to delivery zones or industrial areas in the monitored regions.

\textbf{(4) Taxi Movement is Less Concentrated:}
Unlike private cars, taxis do not have high-density peaks. Instead, they are evenly distributed with mild intensity at multiple locations.
This aligns with expectations, as taxis are constantly moving and less likely to form congestion hotspots compared to private cars.

The heatmap effectively visualizes urban congestion points, thereby supporting traffic flow monitoring and helping identify critical areas for traffic management. 
Additionally, some mild-intensity regions of private cars and pickups overlap, suggesting potential classification challenges due to misclassifications. 
Future enhancements may include implementing time-series traffic modeling to observe peak congestion trends and exploring CNN-based classification improvements for better vehicle differentiation.

\subsubsection{Gradient Correlation Heatmap of Traffic Data Analysis}

\begin{figure}[ht]
\centering
\includegraphics[width=\linewidth]{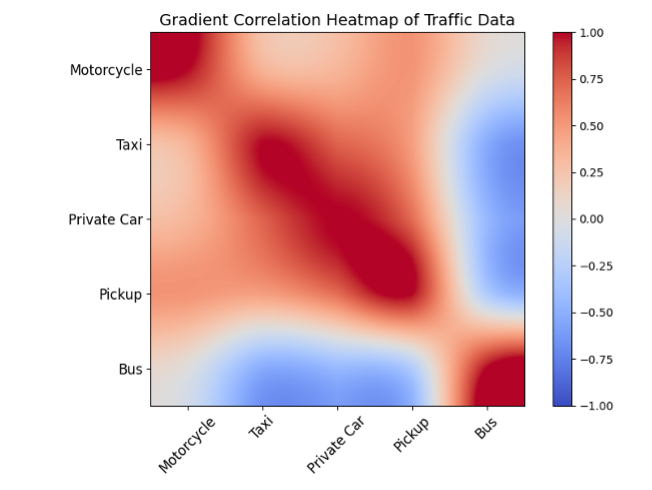}
\caption{Correlation heatmap of vehicle types}
\label{fig7:Correlation}
\end{figure}

Figure~\ref{fig7:Correlation} visualizes the pairwise correlation between vehicle categories based on their total counts at each observation point.

\textbf{(1) Strong Positive Correlations (Red Regions):}
Private cars and taxis exhibit high correlation, meaning areas with high taxi density also tend to have significant private car traffic. 
Motorcycles and pickups show a similar trend, indicating that these two vehicle types may share common traffic patterns, possibly due to commercial and delivery-related movements.

\textbf{(2) Negative Correlations (Blue Regions):}
Buses show negative correlations with private cars and taxis, suggesting that bus-heavy areas experience lower private vehicle traffic, possibly due to efficient public transport infrastructure or dedicated bus lanes. 
Additionally, a slight negative trend is observed between private cars and pickups, indicating that in areas with more private cars, pickups are less common—possibly due to zoning differences such as commercial versus residential land use.

\textbf{(3) Neutral or Weak Correlations (White Regions):}
Some intersections display neutral or weak correlations, meaning that the presence of one vehicle type does not significantly impact the presence of another in certain areas.

These correlation patterns offer valuable insights for traffic demand modeling by highlighting where different vehicle categories tend to co-occur or avoid each other. 
For example, a strong positive correlation between private cars and taxis across several points may indicate zones of high passenger activity, supporting the case for designated taxi stands or ride-hailing areas. 
Conversely, a negative correlation between buses and private vehicles could reveal where public transportation effectively absorbs demand, offering opportunities to expand or replicate such infrastructure. 
These insights can guide urban planning strategies, such as adjusting signal timings, allocating road space, or prioritizing investment in specific transit corridors.
A time-based correlation analysis could reveal peak-hour dependencies by dynamically visualizing correlations at different times of day. 
Additionally, integrating the heatmap data with GIS-based spatial mapping tools could significantly enhance urban traffic planning strategies.

\section{Conclusions} 
\label{sec:4}

This study presents a modular and scalable UAV-based traffic surveillance system capable of conducting vehicle detection, classification, tracking, and behavioral analysis under unconstrained, real-world urban conditions. 
By integrating multi-scale and multi-angle template matching, Kalman filtering, and homography-based spatial calibration, the system achieves robust performance with a detection accuracy of 90.8 \%, classification F1-score of 90.3 \%, and tracking metrics (MOTA/MOTP) exceeding 92 \%.
Applied across diverse traffic scenes and observation points, the system reliably identifies five vehicle categories: motorcycles, taxis, private cars, pickups, and buses, while uncovering dominant traffic patterns such as the consistent prevalence of private cars. 
This result not only validates the classification model under realistic flight altitudes (200 m), but also affirms the system’s suitability for wide-area traffic surveillance.
The calibrated speed estimation pipeline provides localized velocity profiles, revealing expected behaviors such as deceleration near U-turns and congestion buildup at intersections. 
These findings are supported by structured spatial visualizations, including volume heatmaps and inter-class correlation matrices.
The proposed system further demonstrates the ability to detect high-impact driving violations, including: unsafe lane changes (16.3 \% of vehicles in the target zone), illegal double parking (contributing to 11–18 \% delays during peak hours), and crosswalk obstructions (with 42.7 \% of cases involving taxis or private cars).
Such violations are identified through a combination of geofencing, temporal motion filtering, and trajectory deviation analysis enabling precise and context-aware detection.
In addition, traffic count analysis across spatially distributed zones highlights consistent inter-class relationships and congestion-prone locations. 
These results will inform the development of real-time congestion prediction models and adaptive traffic control strategies.
Crucially, unlike prior works limited to simulations or controlled deployments, this framework is validated on real, unstructured aerial footage without reliance on fixed infrastructure, annotated datasets, or privileged sensor placement. 
This ensures high generalizability, cost-effectiveness, and practical relevance to smart city implementations.

While the proposed traffic surveillance system performs reliably across all tasks, three primary limitations are observed:
(1) Reduced recalls in dense occlusion scenarios;
(2) Misclassifications between visually similar classes (pickup vs. private car);
(3) Tracking drift in high-speed vehicles (>100 km/h).
To address these challenges, future work will pursue the integration of: (1) Hybrid convolutional neural network-template detection pipelines to improve occlusion resilience; (2) Advanced vehicle re-identification for long-term tracking continuity; (3) Learning-based speed estimation for extreme motion conditions; (4) Real-time alert systems integrated with UAV software development kits for immediate enforcement response.

In summary, the proposed platform bridges low-level visual analytics with high-level behavioral understanding, enabling a data-driven, enforcement-aware, and infrastructure-independent approach to modern traffic intelligence. 
It provides both a practical toolkit for current urban monitoring needs and a scalable foundation for next-generation intelligent transportation systems.


\bibliographystyle{IEEEtran}
\bibliography{root}

\end{document}